\pdfoutput=1
\documentclass{article}
\PassOptionsToPackage{numbers, compress}{natbib}
\usepackage[final]{neurips_2023}

\usepackage[utf8]{inputenc} 
\usepackage[T1]{fontenc}    
\usepackage{hyperref}       
\usepackage{url}            
\usepackage{booktabs}       
\usepackage{nicefrac}       
\usepackage{microtype}      
\usepackage{xcolor}         

\usepackage{latexsym}
\usepackage{inconsolata}
\usepackage{multirow}

\usepackage{comment}
\usepackage{amsmath,amssymb,amsfonts}
\newtheorem{proposition}{Proposition}
\newenvironment{proof}{\paragraph{Proof:}}{\hfill$\square$}
\usepackage[ruled]{algorithm}
\usepackage{algpseudocode}
\usepackage{caption}
\usepackage{xfrac}
\algdef{SE}[DOWHILE]{Do}{doWhile}{\algorithmicdo}[1]{\algorithmicwhile\ #1}%
\NewDocumentCommand{\LeftComment}{s m}{%
  \Statex \IfBooleanF{#1}{\hspace*{\ALG@thistlm}}\(\triangleright\) #2}
\usepackage{comment}

\usepackage[capitalize]{cleveref}
\crefname{section}{Sec.}{Secs.}
\Crefname{section}{Section}{Sections}
\Crefname{table}{Table}{Tables}
\crefname{table}{Tab.}{Tabs.}
\let\oldnl\nl
\newcommand{\nonl}{\renewcommand{\nl}{\let\nl\oldnl}}
\newcommand{\bs}{\boldsymbol}
\newcommand{\BEA}{\begin{eqnarray}}
\newcommand{\EEA}{\end{eqnarray}}
\newcommand{\argmin}{\mathop{\mathrm{argmin}}}

\def \E {\mathbb{E}} 
\def \RR{\mathbb{R}} 
\def \X{\mathcal{X}} 
\def \Y{\mathcal{Y}} 
\def \D {\mathcal{D}} 
\def \B {\mathcal{B}} 
\def \model {\mathrm{f}} 
\def \config {\Phi} 
\def \configspace {\mathcal{A}} 
\def \wproj {\bs{\Pi}}
\def \supernet {\texttt{Sup}} 
\def \wt {\mathrm{W}} 
\def \wsup {\wt_{\!\supernet}} 
\def \L {\mathcal{L}} 
\def \Lsup {\L_{\supernet}} 
\def \loss {\ell} 
\def \aloss {\texttt{loss}} 
\def \ce {\texttt{CE}} 
\def \kl {\texttt{KL}} 
\def \fproj {\mathbf{U}}
\newcommand{\mse}[2]{\|{#1} - {#2}\|_2^2} 
\def \taskloss {\loss_{\texttt{task}}} 
\def \kdloss {\loss_{\texttt{KD}}} 
\def \fdloss {\loss_{\texttt{FD}}} 
\def \teacher {\texttt{Tch}} 
\def \student {\texttt{Std}} 
\def \pretrain {\mathrm{pretrain}} 
\def \fine-tuned {\mathrm{FT}} 

\def \xx {x} 
\def \yy {y} 
\def \logit {z} 
\def \feature {\mathbf{h}} 
\def \softmax {\bs{\sigma}} 
\def \grad {\texttt{grad}} 

\def \nparam {n} 
\def \dlow {d_{\mathrm{low}}} 
\def \minnet {\mathrm{min}} 

\title{ 
 Efficiently Distilling LLMs for Edge Applications 
}

\author{
Achintya Kundu \\
IBM Research \\
\texttt{ achintya.k@ibm.com } 
\\ \And
Fabian Lim  \\
IBM Research \\
\texttt{ flim@sg.ibm.com }
\\ \And
Aaron Chew \\
IBM Research \\
\texttt{ aaron.chew1@ibm.com }
\\ \AND
Laura Wynter \\
IBM Research \\
\texttt{ lwynter@sg.ibm.com }
\\ \And
Penny Chong  \\
IBM Research \\
\texttt{ penny.chong@ibm.com }
\\ \And
Rhui Dih Lee \\
IBM Research \\
\texttt{ rhui.dih.lee@ibm.com }
}

\begin{document}
\maketitle

\begin{abstract}
Supernet training of LLMs is of great interest in industrial applications as it confers the ability to produce a palette of smaller models at constant cost, regardless of the number of models (of different size / latency) produced. We  propose a new method called Multistage Low-rank Fine-tuning of Super-transformers (MLFS) for parameter-efficient supernet training. We show that it is possible to obtain high-quality encoder models that are suitable for commercial edge applications, and that while decoder-only models are resistant to a comparable degree of compression, decoders can be effectively sliced for a significant reduction in training time. 
\end{abstract}



\section{Introduction}
Given their sizes up to billions of parameters, \cite{raffel2020exploring,brown2020language}, it is challenging for enterprises to fine-tune Large Language Models (LLMs), and furthermore they are not suitable for deployment on edge devices with limited memory and computational power. 
We wish to enable LLMs on edge environments for enterprise use cases. This requires the following two capabilities.
(1) Accommodating a variety of edge device hardware: A single fine-tuned model is not optimal across the spectrum of devices. For industrial applications, a palette of fine-tuned LLMs is required for different hardware. (2) Dynamically changing resource levels: At run-time, the available resources on edge devices evolve over time, and appropriate model should be dynamically selected based on the available resources of each device. 

A considerable amount of research has focused on compressing LLMs \cite{zhu2023survey, sanh2019distilbert,mukherjee2020xtremedistil,mukherjee2021xtremedistiltransformers,jiao2019tinybert,hsieh2023distilling}. Methods that train a single small model guided by a large teacher model such as DistilBERT \cite{sanh2019distilbert} and BERT-PKD \cite{sun2019patient}, either achieve limited compression or do not scale to a large number of deployment devices. Supernet training methods \cite{hou2020dynabert,xu2021bert,cai2019once,tofaCNN,lou2021dynamic,jawahar2023mixtureofsupernets} were introduced to address these limitations: multiple smaller subnets within the supernet are trained simultaneously with  weight-sharing. This one-time training approach produces a palette of smaller models, helping mitigate the  computational cost of fine-tuning a model for each deployment scenario. However, the full-parameter supernet training approach is impractical when fine-tuning of an LLM is required for multiple deployment scenarios, limiting its utility for  enterprises.  

Parameter-efficient fine-tuning (PEFT) methods such as Low-Rank Adaptation (LoRA) reduces the number of trainable parameters by allowing only rank-decomposition matrices to be trained while freezing the pre-trained weights of the model. PEFT methods, however, are not applicable to supernet training due to the implications on the weight-shared subnetworks. Our work bridges this gap to enable efficient fine-tuning of LLMs for edge devices. Our contributions are:

\begin{enumerate}
\item We propose a parameter-efficient, distillation-based approach for supernet training of LLMs.     
\item We devise a gradient scaling scheme to improve convergence speed of any form of supernet training. 
\item We demonstrate significant compression of encoder models for edge. We highlight the  limits of comparable compression for decoder models, while demonstrating a huge reduction in the steps needed for convergence.
\end{enumerate}

\section{Related Work}
Classical compression methods have been used for LLMs including  pruning \cite{mccarley2019structured,voita2019analyzing}, low rank approximation \cite{ma2019tensorized,lan2019albert}, and quantization \cite{shen2020q,zafrir2019q8bert,bhandare2019efficient}. Knowledge distillation (KD)  is adopted in BERT-PKD \cite{sun2019patient}, tinyBERT \cite{jiao2019tinybert}, and distilBERT \cite{sanh2019distilbert} and \cite{gu2023knowledge} in MiniLLM to distill knowledge from the layers of a large transformer model to a smaller one. See also the survey \cite{zhu2023survey}. All these existing methods produce a single compressed model, unsuitable for edge scenarios with multiple deployment devices having varying computational capability.

Neural architecture search (NAS)  based on reinforcement learning \cite{zoph2016neural} and evolutionary algorithms \cite{real2019regularized,zhu2019eena} trains  every possible  architecture and is very slow. Weight-sharing NAS was thus developed: in  \citet{guo2020single, cai2018proxylessnas}, the building blocks in the same layer are isolated as all architectures are single paths. Weight-sharing NAS does not scale well to large architecture search spaces, hence, weight-entangled NAS, where subnets with common  parts share weights, was introduced.

For resource-constrained edge deployment,  supernet training  \cite{cai2019once,tofaCNN,chen2021autoformer,xu2021bert,gao2022autobert,dong2022prior} was developed as a mode of jointly training multiple subnetworks (subnets) with entangled weights:  one trains the supernet  only once for all  deployment scenarios. \citet{cai2019once} introduced an elastic convolutional neural network     with "progressive shrinkage",  where  larger subnets are trained first. Recent works have improved  sampling strategies,  e.g. the sandwich rule with in-place distillation \cite{yu2020bignas},  attentive sampling \cite{wang2021attentivenas},  stochastic nature gradient  \cite{zhang2021you}, or  post-training sampling \cite{lou2021dynamic}. 
Our work is related to supernet training for transformer models \cite{hou2020dynabert,zhang2021you,wang2022lighthubert,wang2020hat,chen2021autoformer}. This  gradient scaling technique can be used with any of the above  supernet methods. 

Parameter-efficient fine-tuning (PEFT) has been of great benefit in fine tuning LLMs. BitFit \cite{bitfit2022} updates the bias terms in pre-trained models while freezing the remaining parameters. LoRA \cite{hu2022lora} decomposes attention weight gradients into low-rank matrices to reduce the number of trainable parameters.  AdaLoRA \cite{zhang2023adaptive} and QLoRA \cite{dettmers2023qlora} further improve  LoRA \cite{hu2022lora}. Note  that  PEFT allows  fine-tuning  a base model on a single GPU but does not produce smaller models. None of the PEFT methods can be used for weight-entangled supernet training. 
  

\section{ Solution Design }
For use in enterprise settings, the solution must allow fine-tuning of models on  a small GPU footprint. 
In addition, inference cost  in terms of storage must be minimised. 
We therefore design a solution which does not  store the full size model checkpoint for every downstream task but only the frozen weights of the pre-trained base model and the low rank matrices. 
For inference in commercial edge use cases, we wish to enable  storing the desired models locally for a wide variety of  edge device resource requirements. We thus develop an approach where  storage is minimised,   storing only one base model and as many low rank adapter matrices as there are target model size variations, where  low-rank adapters  are very small.
If the model is stored locally on an edge device, our proposed slicing operation takes place where the supernet fine-tuning is performed and the desired model is downloaded for inference.   The slicing operation takes place for each model size-task combination and each resulting subnet can  be cached for  inference. 

\section{Problem Formulation}
First, we provide notation. Given a transformer model with architectural configuration $\config$ and weights $\wt$, we denote its forward-pass mapping by $\model_{\config} ( \cdot ; \wt ) : \X \to \Y$. We consider the output space $\Y$ to be the set of all non-negative vectors in $\RR^\nu$ with elements summing to $1$, where $\nu$ denotes the number of classes / vocabulary size). With slight abuse of notation, we  write the forward-pass mapping of an input $\xx \in \X$ through a transformer model $\config$ as 
$\hat{\yy}, \logit, \feature  ~=~ \model_{\config} ( \xx ; \wt ), $
where $\hat{\yy} \in \Y$ denotes the predicted probability distribution over the  (class labels) vocabulary, $\logit$ denotes the vector of logits, and $\feature$ represents a tuple of features such as hidden state vectors  and attention values from different transformer layers. Note that $\hat{\yy} = \softmax(\logit)$, where $\softmax$ is the standard soft-max function that maps a vector of logits into a probability vector.
Given a training data set $\D_{train} \subset \X \times \Y$,  model weights $\wt$ are  learnt by minimizing  training loss:
\BEA \label{eq:1model_training_obj}
    \argmin_{ \wt } \Big[ \, \L_{\config}(\wt) \,:=\, \E \! \big[ \, \loss \! \left[ \, \model_{\config} \! \left( \xx; \wt \right), \yy \,\right] \, \big] \, \Big], 
\EEA
where $\E$ denotes expectation over training example $(\xx,\yy)$ drawn uniformly at random from $\D_{train}$ and $\loss$ denotes a loss function. Most commonly, $\loss$ is chosen to be a  task specific loss function, $\taskloss$, such as    cross-entropy  (i.e.,  $\ce[\cdot,\cdot]$) for classification or causal language modeling loss for generative models.

Next, we introduce the super-transformer and related terminologies.
We  define three types of networks  - \textit{Teacher network}, \textit{Super-transformer} (supernet) and \textit{Sub-transformer} (subnet). The teacher  is a fixed network with the same configuration as the pre-trained transformer. A super-transformer is a dynamic model whose architectural dimensions (embedding dimension, number of heads, number of layers, etc.) are configurable at run time. The \textit{maxnet} (resp. \textit{minnet}) is the largest (resp. smallest) network in the super-transformer's architecture space. Weight entanglement (weight-sharing) allows   super-transformer weights to be used across sub-transformers, which are subsets of the super-transformer. Pre-trained transformer weights initialise the super-transformer. 

The dynamic nature of a super-transformer is explicitly specified via a set $\configspace$, called \emph{configuration space}, consisting of architectural configurations of all sub-transformer models under consideration. The definition of a super-transformer also includes how the configuration $\config \in \configspace$ is to be mapped to a unique transformer model $\model_{\config}$. A weight-sharing super-transformer uses a set of shared weights $\wsup$ to define all sub-transformer models' weights. This is done through a weight projection operator $\wproj$ that slices (selects an appropriate subset of) the super-transformer's weights $\wsup$ into weights of a sub-transformer model: 
\BEA
\wt_{\config} := \wproj_{\config} \! \left( \wsup \right) ,\, \forall \config \in \configspace . \label{eq:wproj}
\EEA
The aim of a weight-sharing super-transformer is to simultaneously train all the transformer models $\{ \model_{\config}( \cdot ; \wproj_{\config}(\wt) ) : \X \to \Y \,|\, \config \in \configspace \}$ through the shared weights $\wsup$. A typical training objective for super-transformers is the training loss averaged over all model configurations in $\configspace$: 
\BEA
\argmin_{ \wsup } \! \Big[ \Lsup(\wsup) := \E \big[ \L_{\config} \! \big( \wproj_{\config} \! ( \wsup ) \big)  \big] \Big], \hspace{-3mm} \label{eq:supernet_training_obj} 
\EEA
where $\E$ denotes expectation over model configuration $\config$ drawn uniformly at random from $\configspace$ and $\L_{\config}$, as defined in \eqref{eq:1model_training_obj}, is  averaged training loss for configuration $\config$. Super-transformer weights, $\wsup$, are  learnt  with stochastic gradient (denoted  $\hat{\nabla}$) of the super-transformer's loss $\Lsup$ estimated as 
\BEA 
    \hat{\nabla}_{\wt} \Lsup( \wsup ) \!=\! \displaystyle  \frac{1}{K} \! \sum_{j=1}^K  \! \hat{\nabla}_{\wt} \L_{\config_j} \!\big( \wproj_{\config_j}\! ( \wsup ) \big),   \!\!\! &\label{eq:supernet_grad} \\
    \hat{\nabla}_{\wt}\L_{\config}( \wt_{\!\config} ) \!=\! \displaystyle \frac{1}{|\B|} \! \sum_{i \in \B } \! \nabla_{\wt} \loss \! \left[ \model_{\config} \! \left( \xx^i;  \wt_{\!\config} \right), \yy^i \right] \!, \!\! & \label{eq:subnet_grad}
\EEA
where $\{\config_1, \cdots, \config_K\}$ are $K$ sub-transformer configurations sampled from $\configspace$ to approximate the expectation in \eqref{eq:supernet_training_obj} and $\B$ is a mini-batch of training examples sampled from $\D_{train}$ to approximate the expectation in \eqref{eq:1model_training_obj}. 
Fine-tuning LLM super-transformers is computationally challenging in enterprise use cases as it involves computing gradients of sub-transformers' loss functions with respect to a huge number of parameters. 

\section{Multistage Low-rank Fine-tuning of Super-transformers} 
We therefore developed  Multistage Low-rank Fine-tuning of Super-transformers (MLFS). Given a teacher model with configuration $\config_{\teacher}$ and pre-trained weights $\wt^{\pretrain}_{\teacher}$,
we  assume that its weights  (denoted   $\wt_{\teacher}$) can be fine-tuned on the given task by learning low-rank matrices $A_{0}, B_{0}$ on top of  pre-trained weights $\wt^{\pretrain}_{\teacher}$: 
\begin{eqnarray}
    \wt_{\teacher} := \wt^{\pretrain}_{\teacher} + A_{0} * B_{0}, \label{eq:maxnet_wt}
\end{eqnarray}
where $A_0,B_0$ are of (low) rank $r$.  Note that  pre-trained weights $\wt^{\pretrain}_{\teacher}$ remain unchanged during  super-transformer fine-tuning. The low-rank matrices, $A_{0}$ and $ B_{0}$, are learnt by minimizing the cross-entropy loss of the teacher model $\model_{\config_{\teacher}}(\cdot;\wt_{\teacher}):\X \to \Y$ over the training data set $\D_{train}$. Specifically, we perform $E_0$ epochs of fine-tuning on the teacher to learn $A_{0}, B_{0}$. This is  stage-$0$ of the multistage fine-tuning algorithm. We denote the teacher weights obtained at the end of stage-$0$ by $\wt_{\teacher}$. 
We now define a super-transformer with maxnet configuration the same as the teacher's. Thus  the super-transformer's weights $\wsup$  are of the same size as the teacher weights $\wt_{\teacher}$). To fine-tune the super-transformer weights $\wsup$, in each of the subsequent stages, we freeze $\wt_{\teacher}$ and propose learning two stage-specific low-rank matrices $A_{s}, B_{s}$, of the same rank, $r$, as $A_{0}, B_{0}$,  that are shared across all sub-transformer models in that stage. To be precise, we impose the following structure on the weights of the sub-transformers at stage-$s$:
\begin{eqnarray}
\begin{array}{l}  
    \wsup :=  \wt_{\teacher}  + \sum_{s=1}^2 A_{s} * B_{s} ,  \label{eq:supernet_wt} \\
    \wt_{\config} = \wproj_{\config} ( \wsup ), ~\forall \config \in \configspace. 
\end{array} 
\end{eqnarray}
Stage-$s$ of the fine-tuning process involves learning only the low-rank matrices, $A_{s}$, $B_{s}$, by minimizing the super-transform loss as in \eqref{eq:supernet_training_obj}. In stage-$1$, we sample sub-transformer models by sampling different widths from the super-transformer keeping the depth (number of layers) same as the maxnet. In stage-$2$, we sample sub-transformer models by sampling different widths as well as depths. We always sample the maxnet model from the super-transformer as the $1^{st}$ sub-transformer model, $\config_1$, at every iteration. We call this   {\bf M}ultistage {\bf L}ow-rank {\bf F}ine-tuning of {\bf S}uper-transformers (MLFS) and present it in the Appendix in Algorithm~\ref{alg:tofaFM}.

\begin{proposition}
Let the individually fine-tuned weights of a subnet, $\config$, be expressed as $\wt_{\config} = \wproj_{\config} ( \wt^{\pretrain}_{\teacher} ) + \Delta \wt_{\config}$. Then,  MLFS  has the following structure on $\Delta \wt_{\config}$:
\begin{eqnarray}
\begin{array}{l}
    \Delta \wt_{\config} =  \wproj_{\config} \left( \sum_{s=0}^2 A_s * B_s \right), ~\forall \config \in \configspace,
\end{array}
\end{eqnarray} 
where $\{A_s, B_s \}_{s=0,1,2}$ are low-rank matrices   shared across all sub-transformers $\config \in \configspace$. 
\end{proposition}
To illustrate the computational savings, recall $\wt^{\pretrain}_{\teacher} \in \RR^{d \times d}$, where $d$ is typically of the order $10^4 - 10^6$. For rank $r$ (typically $< 10$) for the low-rank matrices: $A_{s} \in \RR^{d \times r}$, $B_{s} \in \RR^{r \times d}, ~s=0,1,2$, where $r \ll d$. Then, the number of parameters to be learnt in the MLFS approach is $6rd$. In contrast, full fine-tuning requires updating $d^2$ parameters at every iteration.

    \begin{algorithm}[tbh]
    \caption{Multistage Low-rank Fine-tuning of Super-transformers (MLFS) }\label{alg:tofaFM}
    \vspace{5pt}
     \textbf{Input:} Transformer model (teacher) with configuration $\config_{\teacher}$ \& off-the-shelf pre-trained weights $\wt^{\pretrain}_{\teacher}$, model configuration space $\configspace$ consisting of smaller (than $\config_{\teacher}$) transformer architectures of interest, $\D_{train}$: fine-tuning data set for the target task, $r$: rank of the low-rank matrices to be learned and distillation factor $\alpha \in [0,1]$.\\
    \textbf{Loss functions:} Target task loss $\taskloss$, knowledge distillation loss $\kdloss$, feature distillation loss $\fdloss$.\\
    \textbf{Multistage Training:}
    \begin{algorithmic}[1]
    \For{stage $s = 0,1,2$} 
        \State \textbf{Initialize} the low-rank matrices $\{A_{s}, B_{s}\}$ to be learned at stage $s$.
        \For{\texttt{iteration} $ = 1, ...$}
            \State Get a mini-batch $\B$ of training examples: $\{ (\xx^i,\yy^i) \in \D_{train}\,|\,i \in \B \}$. 
            \State Load the super-transformer model with weights ~$\wsup \leftarrow \wt^{\pretrain}_{\teacher} + \sum_{l=0}^s A_{l} * B_{l}$. 
            \State $\configspace_s \!:=\! \{\config_1, \config_2, \cdots \} \!\! \leftarrow \!\! \textbf{\texttt{sample\_sub-transformers}}(\configspace, \mbox{stage}=s)$. [$\config_1:$~ maxnet]
            \For{ each $\config_j \in \configspace_s$ }
                \State Load the sub-transformer model $\config_j$ with weights $\wt_{\config_j} := \wproj_{\config_j}( \wsup )$.  
                \State $\nparam_j := $ \# of fine-tuning weights in model configuration $\config_j$ .
                \State Compute forward-pass on sub-transformer $\config_j$:\, $\hat{\yy}_{j}^i, \logit_{j}^i, \feature_{\config_j}^i \leftarrow \model_{\config_j}\!(\xx^i; \wt_{\config_j} ),\, \forall i \in \B $.                     
                \State In the case of $\config_1$ (maxnet), set the distillation factor $\alpha$ to $0$. 
                \State Find the loss: $\aloss^i_j  \!\leftarrow \!  (1-\alpha) \taskloss [\hat{\yy}_{j}^i, \yy^i] + \alpha \left( \kdloss [ \logit_{j}^i , \logit_{1}^i ] + \fdloss [\feature_{\config_j}^i, \feature_{\config_1}^i] \right), \forall i \in \B $.  
                \State Compute backward-pass on sub-transformer $\config_j$ to find $( \nabla_{\!\!{A_s}} \aloss^i_j, \nabla_{\!\!{B_s}} \aloss^i_j)$.
            \EndFor     
            \State Update $A_{s}, B_{s}$ using the gradients $( \hat{\nabla}_{\!\!{A_s}} \L_{\supernet}, \hat{\nabla}_{\!\!{B_s}} \L_{\supernet} )$ of the super-transformer's loss:
                \BEA
                   \hspace{10mm} \hat{\nabla}_{\!{\wt}} \Lsup \!=\! \displaystyle \frac{1}{ | \configspace_s | } \!\! \sum_{\config_j \in \configspace_s } \!\!\!\! \left( \! \frac{\nparam_1}{\nparam_j} \! \right)^{\!\gamma} \!\!\hat{\nabla}_{\! \wt } \L_{\config_j}, ~\hat{\nabla}_{\!\wt } \L_{\config_j} \!=\! \frac{1}{|\B|} \!\! \sum_{i \in \B} \nabla_{\!{\wt}} \aloss^i_j,\, \forall \, \wt \in \{ A_s, B_s \}. 
                \EEA
        \EndFor     
    \EndFor
    \end{algorithmic}
    \textbf{Output:} $\{A_{s}$, $B_{s}\}_{s=0}^{2}$ and fine-tuned super-transformer weights: $\wsup \!=\! \wt^{\pretrain}_{\teacher} + \sum_{s=0}^2 A_{s} * B_{s}$. 
    \end{algorithm}

\paragraph{Gradient Scaling}

For faster convergence of the smaller sub-transformers within a super- transformer, we propose a novel weighted-combination of the gradients of the sampled sub-transformers.
\begin{proposition} Let $1^{st}$ sampled sub-transformer, $\config_1$, be the maxnet be in every iteration. Then the scaled gradient of the super-transformer training loss, $\Lsup$, in  Algorithm 1 is given by 
\begin{eqnarray}
\begin{array}{l}
        \sum_{j=1}^K ( \nparam_1 / \nparam_j )^{\gamma} \,\nabla_{\wt} \L_{\config_j}, \label{eqn:grad-scaling} 
\end{array}
\end{eqnarray}
where $\nabla_\wt$ denotes gradient w.r.t. only those weights that are being fine-tuned (in this case only the LoRA matrices), $\nparam_j$ denotes the actual number of trainable weights in model configuration $\config_j$ and $\gamma \ge 1$ is a hyper-parameter. 
\end{proposition}

\begin{proof}
Each sub-transformer gradient in (\ref{eqn:grad-scaling}), $\grad^j$, is  scaled by $( \nparam_1 / \nparam_j )$, which is obtained from the relative weighting of the loss functions. Let $\L_j (\wt)$ denote the $j$-th sub-transformer's loss. Using first-order Taylor expansion, we get:
\begin{eqnarray}
\begin{array}{l}
    \L_{\config_j}(\wt+\bs{\delta} ) \approx \L_{\config_j}(\wt) + \langle \nabla_{\wt} \L_{\config_j}(\wt), \bs{\delta}  \rangle,
\end{array} \nonumber
\end{eqnarray}
where $\langle \cdot, \cdot \rangle$ denotes inner (dot) product operation. 
Therefore, the steepest possible decrease in the loss function $\L_{\config_j}$ can be approximated as: 
\begin{eqnarray}
\begin{array}{l}
    \Delta \L_{\config_j} \approx  \|\nabla_{\wt} \L_{\config_j}(\wt)\|_1 \,|\bs{\delta} |_{max} \approx O(\nparam_j) |\bs{\delta} |_{max},
\end{array} \nonumber
\end{eqnarray}
where we approximate the $\| \cdot \|_1$ norm using the zero-th norm, i.e., number of non-zero elements and $\nparam_j$ stands for the actual number of trainable parameters in sub-transformer configuration $\config_j$. Since the decrease in the loss of a sub-transformer model $\config_j$ is approximately proportional to the number of trainable model parameters $(\nparam_j)$, we scale the losses using $(\nparam_1/ \nparam_j)^{\gamma}, \gamma \ge 1$ so that training losses of smaller sub-transformer models converge at a rate similar to that of larger sub-transformer configurations. Recall that $\nparam_1$ is the maximum number of trainable parameters as $1^{st}$ sampled sub-transformer $\config_1$ is always the maxnet.
\end{proof}

\paragraph{Distillation Loss for Super-transformers:}
Knowledge distillation is straightforward in a fixed-network fine-tuning setting. However, it is less so when fine-tuning a supernet, and in particular, fine-tuning a supernet using the proposed multistage LoRA based approach. Specifically, the subnets receive two types of knowledge distillation (KD) from  the teacher: (a) the usual KD loss that utilizes the output logits of the teacher and (b) distillation of features from 
transformer layers \cite{jiao2019tinybert} of the teacher.  

To define the distillation based losses precisely, let the  forward-pass mapping of an input training sample $\xx^i$ through sub-transformer $\config_j$ be $\hat{\yy}_{j}^i, \logit_{j}^i, \feature_{\config_j}^i \leftarrow \model_{\config_j}\!(\xx^i; \wt_{\config_j} )$, where $\feature_{j}^i := (\feature_{j}^{i,1}, \, \ldots, \, \feature_{j}^{i,l},\,\ldots)$ with $\feature_{j}^{i,l}$ denoting the feature vector from $l$-th layer of sub-transformer $\config_j$.
In super-transformers, the model (maxnet) having the largest configuration, $\config_1$, acts as the teacher and knowledge distillation loss for all other sub-transformers w.r.t the teacher is defined as
\BEA \label{eq:kdlossj}
 \kdloss[ \logit_{\config_j}^i , \logit_{\config_1}^i ] = \kl[\, \softmax( \logit_{\config_j}^i / t) , \,\softmax( \logit_{\config_1}^i / t) \,], \, \forall j >1,\nonumber 
\EEA
where $\kl[\cdot, \cdot]$ denotes the standard KL divergence between two probability vectors, and $ t \ge 1$ is a hyper-parameter called the temperature. Let $d_j$ denote the embedding dimension (hidden size) in sub-transformer  $\config_j$. We compute feature based distillation loss by projecting features $ \feature_{\config_j}^{i, l} \in \RR^{ d_j } $ to a low-dimensional space $\RR^{\dlow }$:
\BEA \label{eq:fdlossj}
\fdloss[ \feature_{\config_j}^i , \feature_{\config_1}^i] = \displaystyle \sum_{l} \beta_{j}^{l} \, 
\mse{ \fproj^{l}_{j}\feature_{\config_j}^{i, l} } 
{ \fproj^{l}_{1} \feature_{\config_1}^{i, g_j(l)} } , \nonumber 
\EEA
where $g_j$ maps each layer index of the sub-transformer configuration $\config_j$ to that of the super-transformer ( / maxnet $\config_1$). In this paper, we propose to share the maxnet's feature projection matrices $\{ \fproj^{l}_{1} \in \RR^{ \dlow \times d_1 } \}$ across all sub-transformer models. We do so by slicing the matrices $\{ \fproj^{l}_{1} \}$: 
\BEA
 \fproj^{l}_j ~:=~[ \, \fproj^{g_j(l)}_1 \, ]_{\config_j} \in \RR^{ \dlow \times d_j },
\EEA
where the operation $[ ~ ]_{\config_j}$ selects appropriate subset of columns depending on the configuration $\config_j$. To reduce the number of user-chosen hyper-parameters, we propose the following hyper-parameter sharing:  $\beta^{l}_{j} := \beta_{ g_j(l) } ,\,\forall j, \,l=1,2,\ldots$.
Thus, apart from setting fewer hyper-parameters, one needs to learn only maxnet's feature projection matrices $\{ \fproj^{l}_{1} \,:\,l=1,2,\ldots\}$, making feature distillation in a super-transformer setting computationally efficient. Additionally, we save computation through use of features only from a fixed subset of maxnet layers for distillation across all sub-transformers: i.e., we use the following subset of maxnet layers: $\{ \, g_{\minnet}(l) \,:\, l=1, \ldots,  \,L_{\minnet} \}$, where $L_{\minnet}$ denotes the number of transformer layers in the smallest sub-transformer $\config_{\minnet}$  and $g_{\minnet}$ maps layer indices of $\config_{\minnet}$ to that of maxnet $\config_1$.

\section{Low-rank approach for Distilling an LLM onto a Pre-trained student}  
In this section, we considerthe standard single-stage distillation approach for training a smaller (student) model on a target task with help from a larger fine-tuned (teacher) model. Here, we assume availability of the following: (i) target data set $\D_{train}$ for fine-tuning, (ii) a task specific loss function $\taskloss$, (iii) a teacher model with configuration $\config_{\teacher}$ and weights $\wt_{\teacher}$ are already fine-tuned for the task, (iv) a smaller model than the teacher, called the student model, with configuration $\config_{\student}$ \& pre-trained weights $\wt_{\student}^{\pretrain}$. The goal is to efficiently fine-tune the student model on the target data with knowledge distillation from the larger teacher model. For computational efficiency, we make the following low-rank assumption (as in LoRA approach) on the fine-tuned weights of student model:
$\wt_{\student} := \wt_{\student}^{\pretrain} + A_{1} * B_{1},$  
where $A_{1}, B_{1}$ are LoRA matrices of low rank (typically rank $r$ is chosen between 4 and 16).

\begin{figure*}[ht]
\begin{minipage}{\linewidth}
    \begin{algorithm}[H]
    \caption{Low-rank approach for Distilling an LLM onto a Pre-trained Student} \label{alg:tofaA1}
    \vspace{5pt}
    \textbf{Input:} Larger transformer model (teacher) with configuration $\config_{\teacher}$ \& weights $\wt_{\teacher}$ already fine-tuned for the given target task, student model with configuration $\config_{\student}$ \& pre-trained weights $\wt_{\student}^{\pretrain}$, $\D_{train}$: fine-tuning data set for the target task, $r$: rank of the low-rank matrices to be learned and distillation factor $\alpha \in [0,1]$. \\
    \textbf{Loss functions:} Target task loss $\taskloss$, knowledge distillation loss $\kdloss$, feature distillation loss $\fdloss$. 
    \begin{algorithmic}[1]
        \State \textbf{Initialize} the low-rank matrices $\{A_{1}, B_{1}\}$ for fine-tuning the student model $\model_{\config_{\student}}$ weights:~ $\wt_{\student} := \wt_{\student}^{\pretrain} + A_{1} * B_{1},$ where $\wt_{\student}^{\pretrain}$ is kept frozen and only $\{ A_{1}, B_{1} \}$ are learned. 
        \For{\texttt{iteration} $= 1,2, \ldots$}
            \State Get a mini-batch $\B$ of training examples:~ $\{ (\xx^i,\yy^i) \in \D_{train}\,|\,i \in \B \}$. 
            \State Compute forward-pass on the teacher model:~ $\hat{\yy}_{\teacher}^i, \logit_{\teacher}^i, \feature_{\teacher}^i ~~ \leftarrow ~ \model_{\config_{\teacher}}\!\left(x^i;\wt_{\teacher} \right),\, \forall i \in \B $. 
            \State Compute forward-pass on the student model:~ $\hat{\yy}_{\student}^i, \logit_{\student}^i, \feature_{\student}^i ~~ \leftarrow ~ \model_{\config_{\student}}\!(x^i; \wt_{\student} ),\, \forall i \in \B $. 
            \State Find the loss:\, $\aloss^i \leftarrow  (1-\alpha) \taskloss [ \hat{\yy}^i_{\student}, \yy^i]  + \alpha \left( \kdloss [ \logit_{\student}^i, \logit_{\teacher}^i ]  +  \fdloss [ \feature_{\student}^i, \feature_{\teacher}^i ]\, \right) , \, \forall i \in \B.$
            \State Compute gradients $( \nabla_{\!\!{A_1}} \aloss^i, \nabla_{\!\!{B_1}} \aloss^i) $ using backward-pass on the student model.
            \State Update $A_{1}, B_{1}$ using the gradients ~$\hat{\nabla}_{\!\!{\wt}} \L_{\config_{\student}} = \displaystyle \frac{1}{|\B|}\sum_{i \in \B} \nabla_{\!{\wt}} \aloss^i, ~\forall \wt \, \in \{ A_1, B_1 \} $. 
        \EndFor     
    \end{algorithmic}
    \textbf{Output:} $\{A_{1}, B_{1}\}$ ~and fine-tuned weights of the student model:~ $\wt_{\student} = \wt_{\student}^{\pretrain} + A_{1} * B_{1}$. 
    \end{algorithm}
\end{minipage}
\end{figure*}

Given an input sample $\xx^i$, the knowledge distillation (KD) loss of the student model $\model_{\config_{\student}}$ with respect to the teacher model $\model_{\config_{\student}}$ is given by 
\BEA \label{eq:kdloss}
\kdloss[ \logit_{\student}^i , \logit_{\teacher}^i ] = \kl[\, \softmax( \logit_{\student}^i / t) , \,\softmax( \logit_{\teacher}^i / t) \,], \nonumber
\EEA
where $\sigma(\cdot)$ denotes the soft-max operation, $\kl[\cdot, \cdot]$ denotes the standard KL divergence between two probability vectors, and $ t \ge 1$ is called the KD temperature.

Given input $\xx^i$, let $\feature_{\student}^i := (\feature_{\student}^{i,1}, \, \ldots, \, \feature_{\student}^{i,l},\,\ldots)$ denote the feature vectors from all  transformer layers of the student model with $\feature_{\student}^{i,l}$ being the feature vector from $l$-th layer. Now, we define the feature based distillation loss as 
\BEA \label{eq:fdloss}
\fdloss[ \feature_{\student}^i , \feature_{\teacher}^i] \!=\!\!\! \sum_{l} \beta_{l} \, 
\mse{ \fproj_{\student}^{l} \feature_{\student}^{i, l} }
{ \fproj_{\teacher}^{l} \feature_{\teacher}^{i, g(l)} } , \!\! \nonumber
\EEA
where the summation $l$ is over (a subset of) the layers of the student model, $\{ \beta_l \ge 0 \}$ are user chosen constants, $\|\cdot\|_2$ denotes the Euclidean norm, and the function $g$ maps each layer index $l$ of the student model to a layer index $g(l)$ of the teacher model for comparing corresponding features. The projection matrices $\{ \fproj_{\student}^{l},\fproj_{\teacher}^{l} \}$ are simultaneously learned along with the LoRA weights and are needed to compare the feature vectors from the student with those from the teacher by projecting them to a common low-dimensional space. To be precise, $\feature_{\student}^{i,l} \in \RR^{d_{\student}}$, $\feature_{\teacher}^{i,l} \in \RR^{d_{\teacher}}$, $\fproj_{\student}^{l} \in \RR^{ \dlow \times d_{\student}}$, $\fproj_{\teacher}^{l} \in \RR^{ \dlow  \times d_{\teacher}}$, $\dlow << d_{\student} \le  d_{\teacher}$, where $d_{\teacher}$ \& $d_{\student}$  denote the embedding (hidden) dimension of the teacher and the student model respectively. Usually, the dimension $\dlow$ of the projected space is chosen to be small, 128 in our experiments.


\section{ Results on Encoder and Decoder LLMs}
\label{sec:experiments}
We report performance on encoder   tasks using   GLUE \cite{wang-etal-2018-glue}  with  BERT$_{base}$  as the teacher model  $\config_{\teacher}$. For decoder  LLMs, we use Santacoder  \cite{santacoder} and Codellama7B  \cite{rozière2023code} on a python coding task using  \emph{bigcode/the-stack} data   \cite{bigcodestack}.
We report performance of the sub-transformer models  at the end of stage $s=2$. On GLUE, we use  the train set for fine-tuning and the dev set for accuracy evaluation. For santacoder, we evaluate performance using 
HumanEval \cite{humaneval} and report pass@1 scores.
All  experiments were conducted using PyTorch on a single Nvidia A100 (40GB) GPU. 

In MLFS, the Low rank matrices are added on the QKV vectors and the intermediate size of feed-forward network (FFN) layers.  We set $\beta_l=0.1 \,\forall l$ in feature distillation loss and choose distillation factor $\alpha =0.9$. For training, we use a maximum sequence length of 128; effective batch size of 128 for QQP, MNLI, QNLI, and 64 for the other data sets. Training is done for a maximum of 8 epochs for all GLUE data sets except SST-2 for which we allocate maximum 3 epochs. We set an initial learning rate of $5e^{-4}$ for QNLI \& MNLI, and $1e^{-3}$ for other GLUE data sets. We use rank $r=8$ for the low rank matrices $A, B$ unless mentioned otherwise. We choose gradient scaling hyper-parameter $\gamma=1$ for SST-2 and $\gamma=2$ for all other data sets.
Following \cite{hu2022lora}, we use the fine-tuned MNLI checkpoint to initialize the model weights for experiments on small data sets such as RTE and MRPC.

\subsection{Performance of Encoder Models}

\begin{figure}[htb]
\begin{center}
\includegraphics[width=0.95\textwidth]{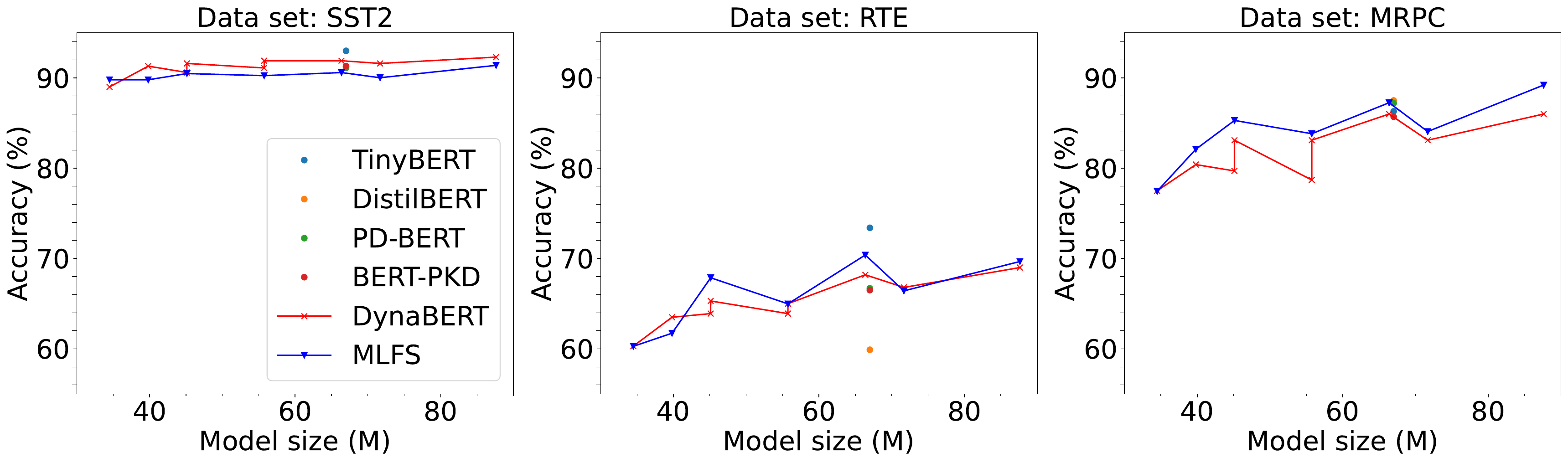}
\vspace{1mm} \\
\includegraphics[width=0.95\textwidth]{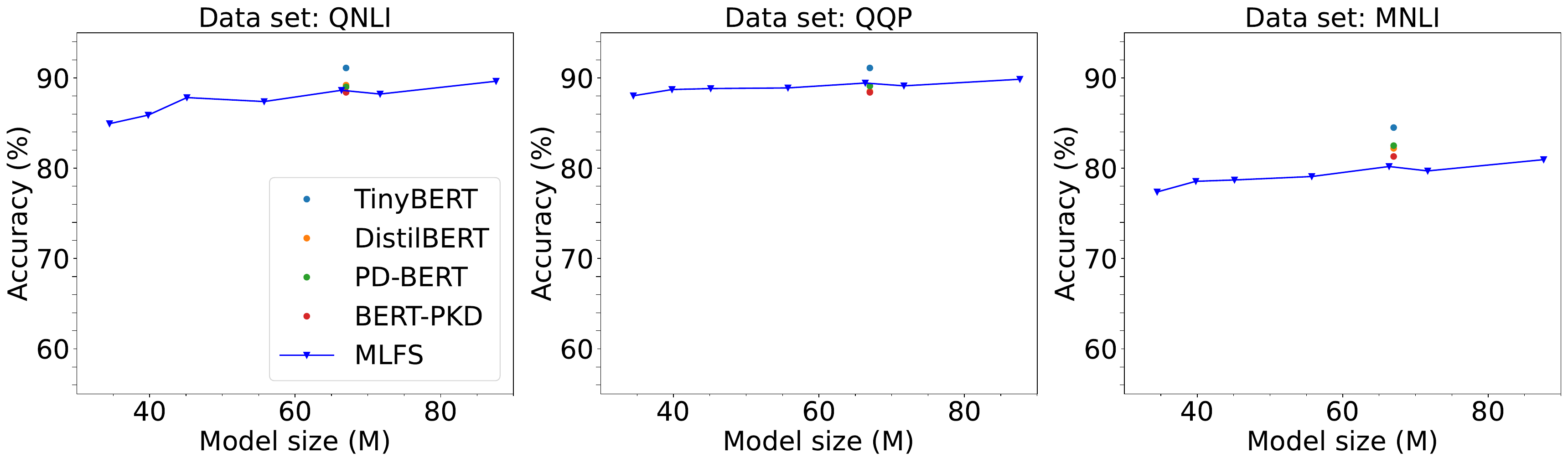} 
\caption{Model size vs performance trade-off for task-specific BERT models produced by MLFS against other methods on 6 GLUE data sets. \\}
\label{fig:glue_results_paramsize}
\end{center} 
\end{figure}

\begin{figure}[bht]
\begin{center}
\includegraphics[width=0.95\textwidth]{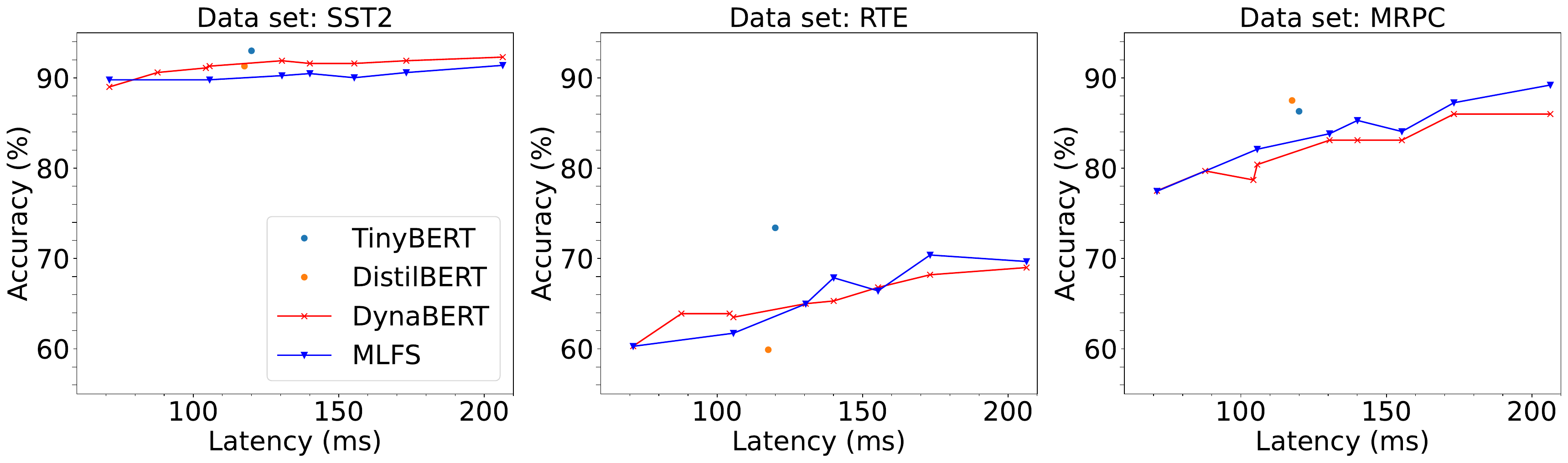}
\vspace{1mm} \\
\includegraphics[width=0.95\textwidth]{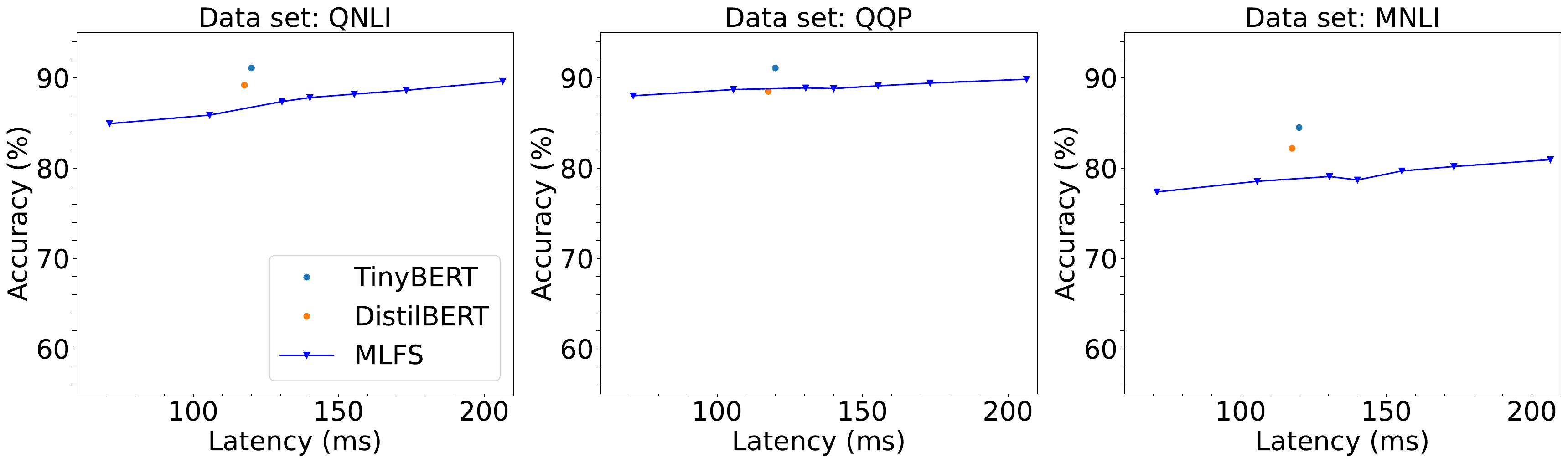} 
\caption{Latency vs performance trade-off for task-specific BERT models produced by MLFS against other methods on 6 GLUE data sets.}
\label{fig:glue_results_latency}
\end{center}
\end{figure}

We compare performance of encoder models obtained with the MLFS approach against a static, fixed model (BERT base) from \cite{zhang2021you,hou2020dynabert}, two popular distilled variants of the fixed model: TinyBERT \cite{jiao2019tinybert} and DistilBERT \cite{sanh2019distilbert}, and models trained using existing super-transformer  methods (DynaBERT \cite{hou2020dynabert}.   
Figure \ref{fig:glue_results_paramsize} shows the performance of the palette of models,  from a  45M param. minnet  to  full-size 110M maxnet. Model performance against forward-pass latency is plotted in Figure \ref{fig:glue_results_latency}. Encoder models produced by MLFS are at par or better than much costlier methods.
Results of PD-BERT, BERT-PKD are from \cite{zhang2021you},    static BERT    from \cite{zhang2021you} for all except MRPC for which we use \cite{hou2020dynabert}.
Note that  TinyBERT 
performs  data augmentation leading to  higher accuracy  but much longer computation time. We do not perform data augmentation for fairness of the comparison to the other methods. 
The main observation is that MLFS  provides accurate, smaller encoder models at 1/4  the size of the teacher  and 1/3 its runtime latency on a single GPU.

\begin{figure}[bht]
\hspace{1.7cm} \includegraphics[width=0.32\textwidth]{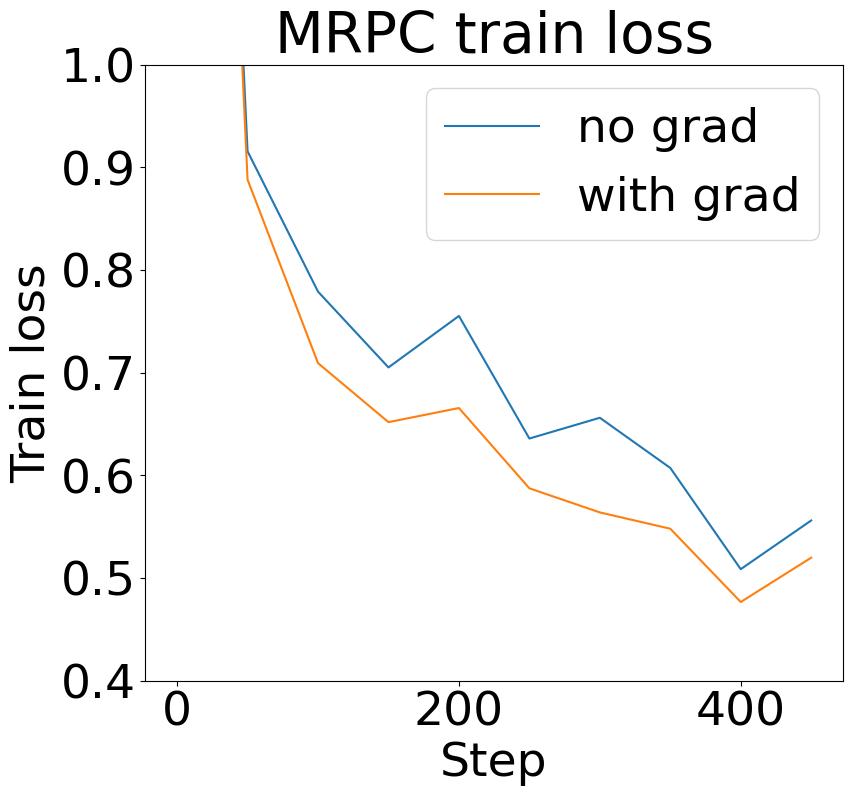} 
\hspace{1.2cm} \includegraphics[width=0.32 \textwidth]{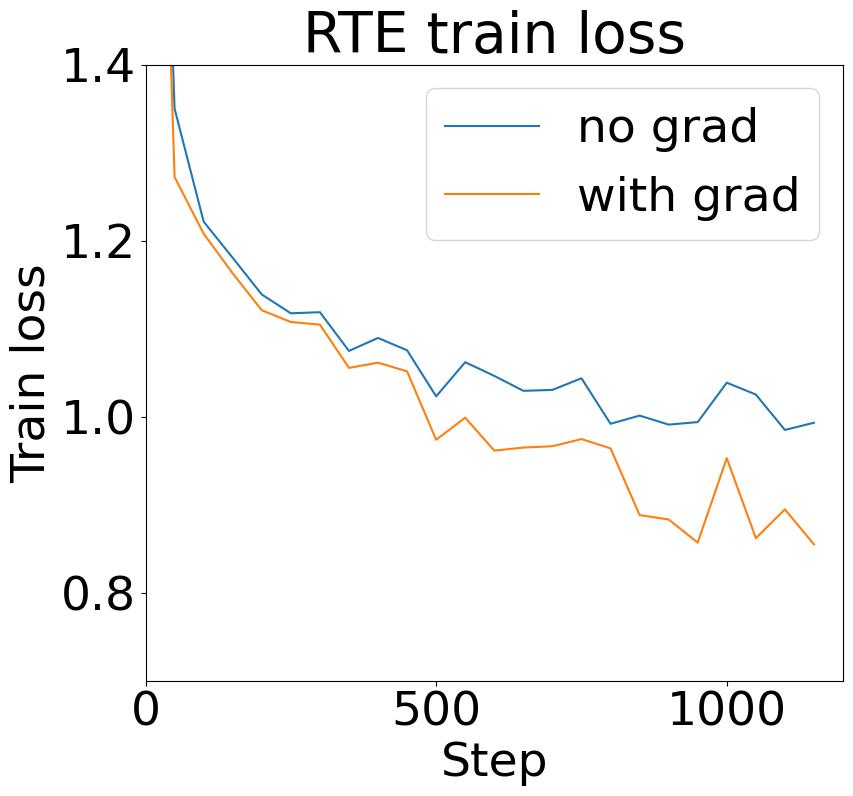} 
\caption{Ablation study on gradient scaling: MLFS minnet  convergence is improved using gradient scaling.}
\label{fig:ablation_gradscaling_trainingloss}
\end{figure}

\begin{figure}[ht] 
\centering
\includegraphics[width=0.7\columnwidth]{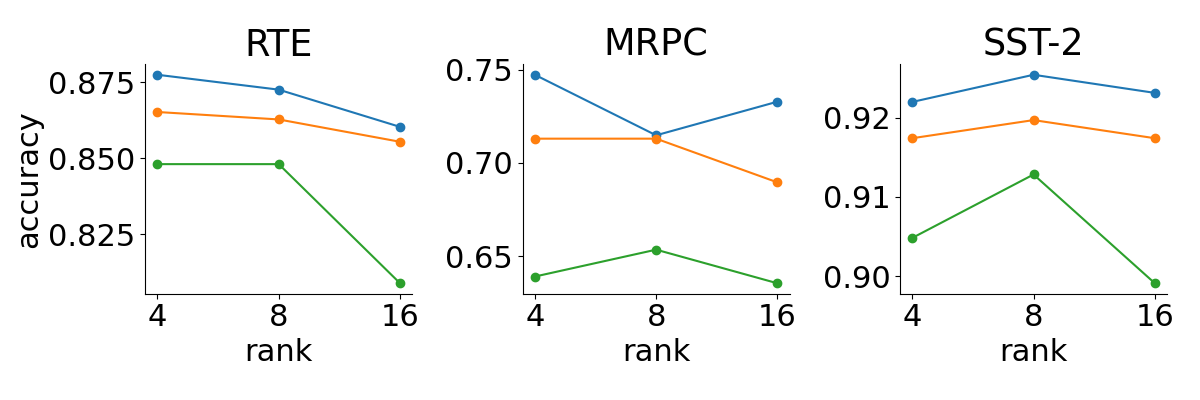}
\caption{Ablation study on MLFS rank of $A,B$.  Maxnet (top: blue),  minnet (bottom: green), and average of two medium-sized subnets (middle: orange). Rank $r=8$ is optimal for small and medium subnets.}
\label{fig:ablation_lora_r}
\end{figure}

\paragraph{Ablation Study on Gradient Scaling}
In supernet training, the weights of maxnet and subnets are shared and trained simultaneously. The maxnet tends to converge and overfit earlier than  smaller subnets. The different convergence rates renders selecting a single supernet checkpoint for all networks difficult.  Gradient scaling  solves this by speeding up  convergence of the smaller subnets to match that of the larger subnets or the maxnet. 
Fig. \ref{fig:ablation_gradscaling_trainingloss} shows that  gradient scaling  improves minnet convergence, indicated by  lower minnet loss.

\paragraph{Ablation Study on Rank of $A,B$}
In Fig. \ref{fig:ablation_lora_r}, we examines the impact of rank $r$ of the matrices $A,B$ on performance. Note that the actual number of parameters fine-tuned vary as we vary the rank $r$. The aim is to provide good results for the smaller networks. Here, rank $r=8$ works well across the GLUE data sets. Therefore, we use rank $r=8$ for $A, B$ for all other MLFS experiments. From the scale of the y-axis in \ref{fig:ablation_lora_r},   observe that MLFS is not overly sensitive to the chosen rank.

\subsection{Performance of Decoder Models}

\begin{figure}[ht] 
\begin{center}
\includegraphics[width=0.8\textwidth]{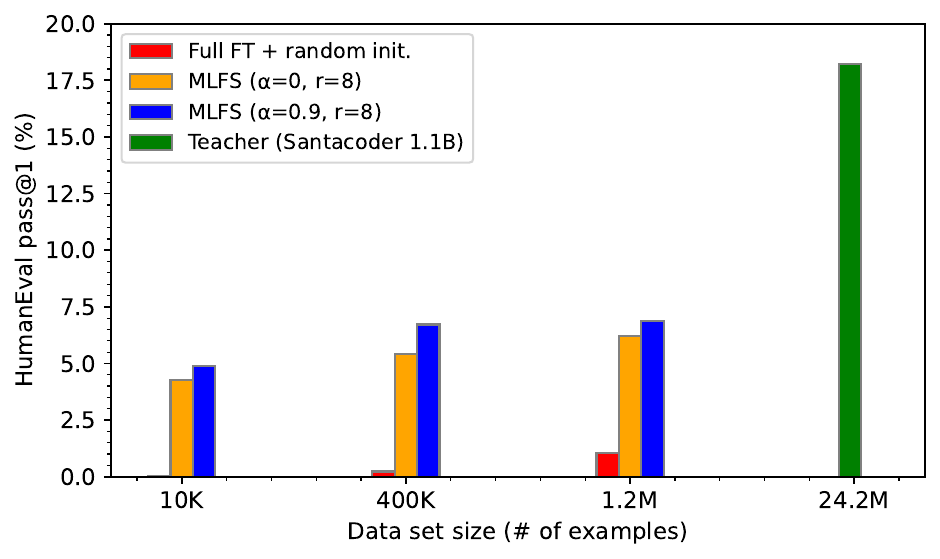} 
\caption{Performance of MLFS on a custom Santacoder 0.7B model using 10K/400K/1.2M training examples.}
\label{fig:santacoder_pass1_MLFS_vs_FT}
\end{center}
\end{figure}

Turning now to decoder models, we consider  two  code-pre-trained LLMs,
Santacoder  \cite{santacoder} and Codellama7B \cite{rozière2023code}. We evaluate a custom 0.7B parameter Santacoder  model obtained from the 1.1B  teacher. Due to an inability to fine-tune on the full  24M  coding examples, we use up to 1.2M. Fig. \ref{fig:santacoder_pass1_MLFS_vs_FT} shows that  MLFS pass@1  improves rapidly as  number of tokens increases from a low 10k  to 400k  to 1.2M examples, only 5\% of the  24M examples.  Table \ref{tab:santacoder_MLFS_palette} shows analogous results with 3 small MLFS models. The improvement in pass@1  indicates that the smaller models retain the ability to learn  from the larger teacher.

\begin{table}
\centering
\begin{tabular}{l|ccc}
 \multirow{2}{*}{Data set size} &
      \multicolumn{3}{c}{Model size}  \\ \cline{2-4}
 & \textbf{0.5B} & \textbf{0.7B} & \textbf{0.9B}\\ \hline \hline
 10K    & 4.5 &   8.6  & 13.4 \\
 400K   & 4.7 &   9.5 & 13.5  \\ \hline
\end{tabular}
\caption{HumanEval pass\texttt{@}1\,(\%) performance of 3 small models produced by MLFS from Santacoder 1.1B.}
\label{tab:santacoder_MLFS_palette}
\end{table}

Contrary to   encoder models, the compression levels that  retain sufficient performance of the teacher with decoders is less. While MLFS  retains  accuracy performance of encoder models at 1/4  the size of the teacher,  the decoder models are reduced to at most  2/3  the teacher's size.

\begin{figure}[ht]
\begin{center}
 \includegraphics[width=.8\textwidth]{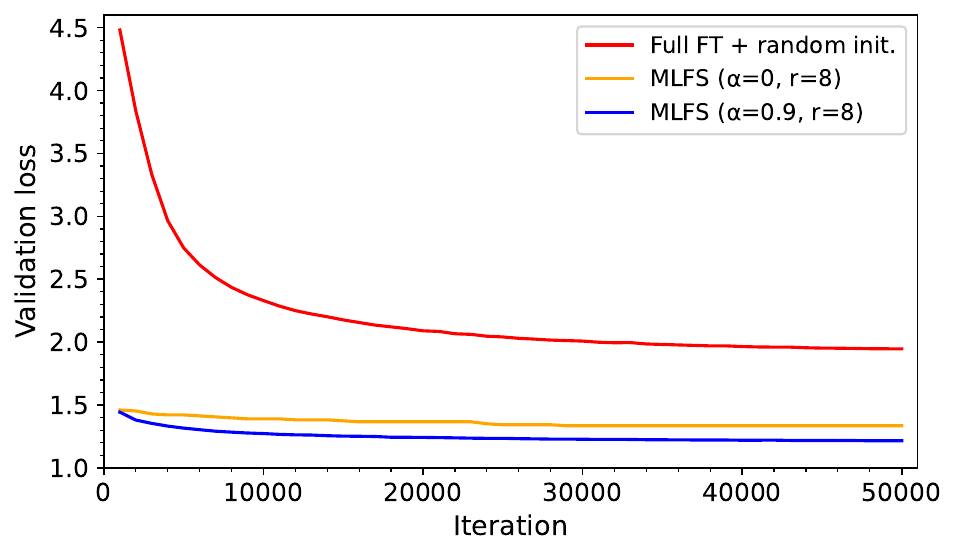}
\caption{Convergence comparison of validation loss while fine-tuning a custom model from random vs using MLFS. MLFS achieves low validation loss much faster.}
\label{fig:santacoder_valloss}
\end{center}
\end{figure}

MLFS slicing of the teacher model can, however, benefit decoder models by reducing substantially the training/fine-tuning time needed  compared to a randomly-initialised model, as shown in Fig. \ref{fig:santacoder_valloss} on Santacoder sliced from 1.1B to 0.7B. In other words, when a smaller model is required for edge inference, one can  train it from a random initialisation, or slice from a  teacher as does MLFS, and train starting from the sliced weights. The latter significantly reduces training time as seen in the validation loss curves.  We see the same benefit  on Codellama, as shown below. See  \cite{samragh2023weight} for a similar observation.

\paragraph{Results on CodeLlama-7B-Python}
Here, we experiment on fine-tuning custom smaller models of different sizes using the \emph{CodeLlama-7B-Python} model as the teacher. We apply 1 epoch of MLFS on 2 different subsets of the \emph{bigcode/the-stack} data set and produce 3 smaller variants having 4.5B, 5.3B \& 6B model parameters in each case. Table \ref{tab:codeLlama_MLFS_palette} shows HumanEval performance of these models as the number of examples is increased from 200K to 400K. Again,  we see that the sliced CodeLlama  retain their ability to learn and improve quickly as the number of examples increases. Note that the full data set includes 24M examples; MLFS achieves nearly 75\% of the performance of  fullsize CodeLlama after less than 2\% of the examples. 

\begin{table}[ht]
\centering
\begin{tabular}{l|ccc}
 \multirow{2}{*}{Data set size} &
      \multicolumn{3}{c}{Model size}  \\ \cline{2-4}
 & \textbf{4.5B} & \textbf{5.3B} & \textbf{6B}\\ \hline \hline
 200K   & 11.0 &   19.5 & 23.2 \\
 400K   & 14.0 &   28.1  & 30.5  \\ \hline
\end{tabular}
\caption{HumanEval pass\texttt{@}1\,(\%) performance of 3 small models produced by MLFS from CodeLlama-7B-Python}
\label{tab:codeLlama_MLFS_palette}
\end{table}

\begin{figure}[htb] 
\begin{center}
\includegraphics[width=.8\textwidth]{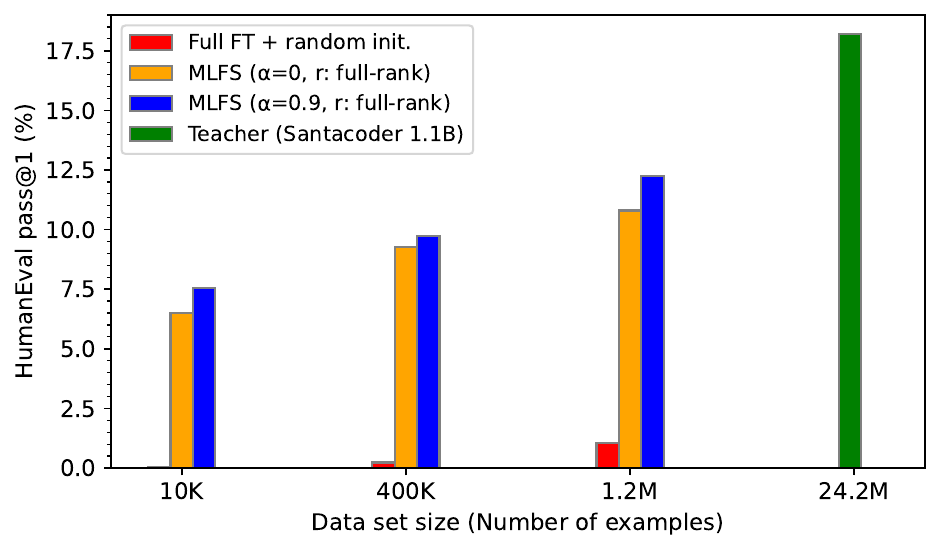} 
\caption{Superior performance of supernet training compared to other full fine-tuning based approaches on three data sets with 10K/400K/1.2M examples.}
\label{fig:santacoder_pass1_Full_FT}
\end{center}
\end{figure}

\paragraph{Ablation Study on Santacoder}
In Fig. \ref{fig:santacoder_pass1_Full_FT}, we compare HumanEval performance of a 0.7B Santacoder model fine-tuned through full fine-tuning (FT) from random initialisation vs. full-rank (non-LoRA) MLFS with ($\alpha=0.9$) and without ($\alpha=0$) distillation. The improvement in the evaluation numbers is remarkable even after fine-tuning on up to only 5\% of the examples. 

\begin{figure}[htb]
\begin{center}
 \includegraphics[width=.8\textwidth]{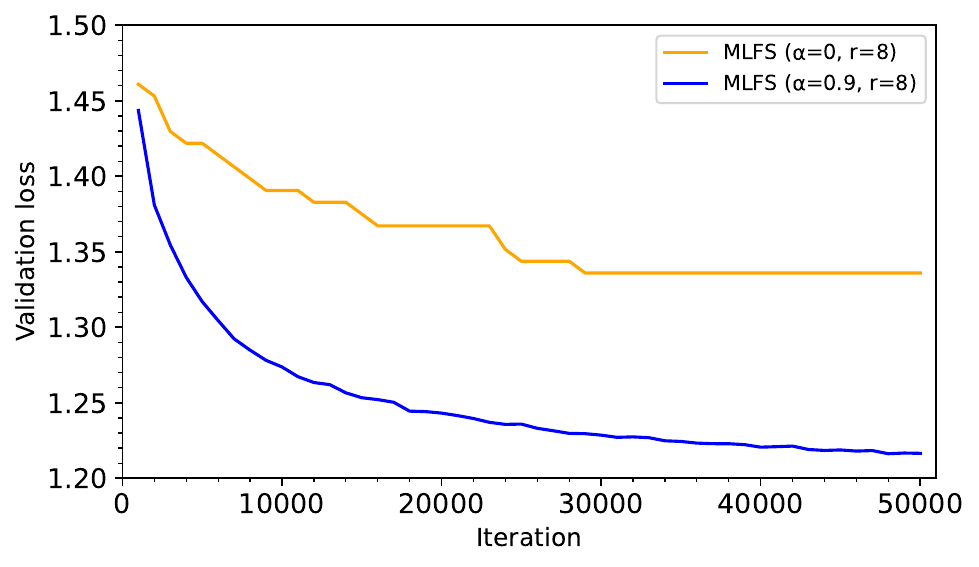}
\caption{Convergence comparison of validation loss while fine-tuning a custom model using MLFS with/without distillation.}
\label{fig:santacoder_valloss_only_mlfs}
\end{center}
\end{figure}

In Fig.~\ref{fig:santacoder_valloss_only_mlfs}, we see better convergence of validation loss on the Santacoder 0.7B for MLFS with distillation loss ($\alpha >0$). This demonstrates the benefit of MLFS distillation as compared to full MLFS fine tuning of the  sliced model.

\section{ Perspectives } 
Enterprise users require an efficient way to fine-tune LLMs for inference on  edge devices of many sizes.  We developed MLFS  for such edge deployment scenarios. We demonstrate its benefits on encoder LLMs. We  show the limitation of compressing decoder LLMs to a comparable degree; however, MLFS offers significant gains for smaller decoder  training/fine-tuning by slicing from a larger pre-trained teacher.

\bibliographystyle{plainnat}
\bibliography{tofafm}

\end{document}